\documentclass{uai2023} 

\usepackage[american]{babel}

\usepackage{natbib} 
\usepackage{mathtools} 
\usepackage{siunitx} 
\usepackage{booktabs} 
\usepackage{amssymb}
\usepackage{subcaption}
\usepackage{listings}
\newcommand{\p}[1]{\left( {#1} \right)}
\newcommand{\s}[1]{\left[ {#1} \right]}
\newcommand{\br}[1]{\left\{ {#1} \right\}}
\newcommand{\norm}[1]{\left|\left| {#1} \right|\right|}

\title{Anomaly Detection via Gumbel Noise Score Matching}

\author[1]{\href{mailto:<amahmood@cs.unc.edu>?Subject=Your UAI 2023 paper}{Ahsan~Mahmood}{}}
\author[1]{Junier~Oliva}
\author[1]{Martin~Styner}
\affil[1]{%
    Department of Computer Science \\
    University of North Carolina at Chapel Hill
}

\begin{document}
\maketitle

\begin{abstract}
We propose Gumbel Noise Score Matching (GNSM), a novel unsupervised method to detect anomalies in categorical data. GNSM accomplishes this by estimating the scores, i.e. the gradients of log likelihoods w.r.t.~inputs, of continuously relaxed categorical distributions. We test our method on a suite of anomaly detection tabular datasets. GNSM achieves a consistently high performance across all experiments. 

We further demonstrate the flexibility of GNSM by applying it to image data where the model is tasked to detect poor segmentation predictions. 
Images ranked anomalous by GNSM show clear segmentation failures, with the outputs of GNSM strongly correlating with segmentation metrics computed on ground-truth. 
We outline the score matching training objective utilized by GNSM and provide an open-source implementation of our work.

\end{abstract}

\section{Introduction}\label{sec:intro}

Anomaly detection on tabular data remains an unsolved problem~\cite{pang_deep_2021, ruff_unifying_2021, aggarwal_introduction_2017}. Notably, there are few methods in this space that explicitly model categorical data types (~\cite{pang2021homophily}). For instance, none of the methods tested in the recent comprehensive benchmark performed by~\cite{han2022adbench} make explicit use of categorical information. After transforming the categorical variables into one-hot and binary encodings, existing methods proceed to treat them as distinct continuous variables. Furthermore, there is a dearth of unsupervised deep learning anomaly detection methods that excel on tabular datasets. For example, the otherwise exhaustive benchmark of~\cite{han2022adbench} reports only two unsupervised deep learning models ( DSVDD \cite{pmlr-v80-ruff18a} and DAGMM \cite{zong2018deep} ) in their analysis; with both models being outperformed by shallow unsupervised methods. Some reconstruction-based autoencoder approaches have been proposed~\cite{hawkins2002outlier} but they require optimization tricks such as adaptive sampling, pretraining, and ensembling to work effectively~\cite{chen2017outlier}.

To fill this gap, we propose a novel unsupervised method to detect anomalies: Gumbel Noise Score Matching (GNSM). Our method estimates the scores of continuous relaxations of categorical variables.
Note that each relaxed categorical will be represented by the probability vector of all outcomes pertaining to that feature.
It follows that the score (gradients of log likelihood) of a relaxed category will be defined over all outcomes. Thus, our model is implicitly aware of the distribution of outcomes for each categorical feature. We postulate that modeling this information will help our model identify samples composed of rarely occurring categories i.e. anomalies.

Our main contributions are :

\begin{itemize}
    \item Deriving an unsupervised training objective for learning the scores of categorical distributions
    \item Demonstrating the capability of score matching for anomaly detection on categorical types in both tabular and image datasets
    \item Providing a unified framework for modeling mixed data types via score matching
\end{itemize} 

To illustrate the significance of our last contribution, consider the Census dataset in our experiments (Section~\ref{sec:experiments}). We were able to compute the scores for both the continuous features (using standard denoising score matching~\cite{Vincent2011}) and the categorical features (using GNSM). Further still, our model is not limited to tabular data. As demonstrated in Section~\ref{seg_results}, GNSM can effectively detect anomalies in images (segmentation masks). This flexibility, paired with our simple loss objective, illustrates the practical viability of our method.

\section{Background}
Our work combines continuous relaxations for categorical data~\cite{jang2017categorical},\cite{maddison2017concrete} into the denoising score matching objective~\cite{Vincent2011}. We will briefly expand on some of the background material to provide context.

\subsection{Score Matching}

\cite{Hyvarinen2005} introduced score matching as a methodology to estimate the gradient of the log density with respect to the data (i.e. the score): $\nabla_x \log p(x)$. If we assume a noise distribution $q_{\sigma}(\tilde{x}|x)$ is available, it is possible to learn the scores for the perturbed data distribution $q_{\sigma}(\tilde{x}) \triangleq \int q_{\sigma}(\tilde{x}|x) p(x) dx $. \cite{Vincent2011} proved that that minimizing the Denoising Score Matching (DSM) objective in Equation ~\eqref{eq:dsm} will train the score estimator $s_\theta$ to satisfy $s_\theta(x) = \nabla_x \log q_\sigma(x)$.

\begin{equation}
\label{eq:dsm}
    J_{\text{DSM}}(\theta) = \mathbb{E}_{q_{\sigma}} \s{ || s_\theta(x) - \nabla_{\Tilde{x}} \log q_{\sigma}(\tilde{x} | x) ||^2 }
\end{equation}

\cite{song2019generative} introduced Noise Conditioned Score Networks (NCSN) and expanded the DSM objective in~\eqref{eq:dsm} to include multiple noise distributions of increasing noise levels. 

\begin{equation}
\label{eq:ncsn}
    J_{\text{NCSN}}(\theta) = \sum_{i=1}^L \mathbb{E}_{q_{\sigma_i}} \s{ || s_\theta(x, \sigma_{i}) - \nabla_{\Tilde{x}} \log q_{\sigma_{i}}(\tilde{x} | x) ||^2 }
\end{equation}

The authors' main insight was to use the same model for all noise levels. They parameterized the network to accept noise scales as conditioning information. NCSNs were successful in generating images and have been shown to have close ties to generative diffusion models~\cite{song2019generative}.

\subsection{Connecting Score Matching to Anomaly Detection }

While~\cite{song2019generative} demonstrated the generative capabilities of NCSNs, ~\cite{mahmood2021multiscale} outlined how 
these networks can be repurposed for outlier detection. Their methodology, Multiscale Score Matching Analysis (MSMA), incorporates noisy score estimators to separate in- and out-of-distribution (OOD) points. Recall that a score is the gradient of the likelihood. A typical point, residing in a space of high probability density will need to take a small gradient step in order to improve its likelihood. Conversely, a point further away from the typical region (an outlier) will need to take a comparatively larger gradient step towards the high density region.  When we have multiple noisy score estimates, it is difficult to know apriori which noise scale accurately represents the gradient of the outliers. However, ~\cite{mahmood2021multiscale}, showed that learning the typical space of score-norms for all noise levels is sufficient to identify anomalies.

Concretely, assume we have a score estimator that is trained on $L$ noise levels and a set of inlier samples $X_{\text{IN}}$. Computing the inlier score estimates for all noise levels and taking the L2-norms across the input dimensions results in an $L$-dimensional feature vector for each input sample: $[\norm{  s (X_{\text{IN}}, \sigma_1) }_2^2, ... , \norm{  s (X_{\text{IN}}, \sigma_L) }_2^2 ]$. The authors in~\cite{mahmood2021multiscale} argue that inliers tend to concentrate in this multiscale score-norm embedding space. It follows that one could train an auxiliary model (such as a clustering model or a density estimator) to learn this score-norm space of inliers. At test time, the output of the auxiliary model (e.g. likelihoods in the case of density estimators) is used as an anomaly score. Results in~\cite{mahmood2021multiscale} show MSMA to be effective at identifying OOD samples in image datasets (e.g. CIFAR-10 as inliers SVHN as OOD). 

\subsection{Continuous Relaxation to Categorical Data}

Note that gradients of log likelihoods are not defined for categorical inputs. In order to compute the score of categorical data, we propose to adopt a continuous relaxation for discrete random variables co-discovered by~\cite{jang2017categorical, maddison2017concrete}. These relaxations build on the Gumbel-Max trick to sample from a categorical distribution ~\cite{maddison2014sampling}. The procedure (often referred to as the Gumbel-Softmax) works by adding Gumbel noise~\cite{gumbel1954statistical} to the (log) probabilities and then passing the resulting vector through a softmax to retrieve a sharpened probability distribution over the categorical outcomes. Of particular interest to us, this procedure incorporates a temperature parameter ($\lambda$ in Equation~\eqref{eq:log_concrete}) to control the sharpening of the resulting probabilities. We argue that this temperature can also be interpreted as a noise parameter, by virtue of it increasing the entropy of the post-softmax probabilities. We will make use of this intuition when we combine continuous relaxations with denoising score matching in Section~\ref{combine}. 

For our work, we will be utilizing the formulation of \cite{maddison2017concrete} i.e. Concrete random variables. In particular, we will be using ExpConcrete Distribution as it is defined in the log domain. This is mainly done for convenience and numerical stability. Given unnormalized probabilities $\alpha \in (0, \infty)^K$, Gumbel i.i.d samples $\mathit{G}_k$, and a smoothing factor $\lambda \in (0, \infty)$ a we can construct a $Y \in \mathbb{R}^n$ such that $\exp(Y) \sim \text{Concrete}(\alpha, \lambda) $:

\begin{equation}
\label{eq:log_concrete}
    Y_k = \frac{\log \alpha_k + G_k}{\lambda} - \log \sum_{i=1}^{K} \exp \br{\frac{\log \alpha_i + G_i}{\lambda}}
\end{equation}

As $\lambda \rightarrow 0$, the computation approaches an argmax, while large values of $\lambda$ will push the random variable towards a uniform distribution.





\section{Score Matching with Categorical Variables}

In this section we will develop the ideas behind our loss objective.

We start by noting that the proof of the denoising objective by \cite{Vincent2011} (Equation~\eqref{eq:dsm}) holds true for any $q_\sigma$, provided that $\log q_{\sigma}(\Tilde{x} |x)$ is differentiable.
Recall that $q_\sigma$ plays the role of a noise distribution. While most denoising score matching models incorporate Gaussian perturbation~\cite{song2019generative,song2021scorebased,Vincent2011}, we emphasize that \textit{any} noise distribution may be used during training.

\subsection{ExpConcrete($\alpha$,~$\lambda$) as a Noise Distribution}
\label{combine}
Following the reasoning above and the temperature parameter ($\lambda$) available in Equation~\ref{eq:log_concrete}, we posit that one can repurpose the Concrete distribution to add `noise' to our continuous relaxations of the categorical variables. Increasing $\lambda$ will allow us to corrupt the input $x$ by scaling the logits and smoothing out the categorical probabilities.

For our objective, we can set the noise distribution as:

\begin{align*}
     \log q_{\sigma}(\Tilde{\mathbf{x}} | \mathbf{x})  &= \log p_{\lambda}(\Tilde{\mathbf{x}}; \alpha=\mathbf{x})
\end{align*}

Here we set the location parameter to that of the unperturbed input (similar to how one would use a Gaussian kernel) with $\lambda$ being a known hyperparameter. Note that $\mathbf{x} \in \br{0,1}^K$ will be a one-hot encoding for $K$ outcomes, which does not satisfy the requirement $\alpha \in (0, \infty)^K$. This can be circumvented by adding a small delta to the vectors to avoid zero values. While it is possible to transform $\mathbf{x}$ to any positive unnormalized probabilities, we opted to use the one-hot encodings for simplicity i.e. $\alpha = \mathbf{x} + \delta $.

\subsection{Denoising Objective}
Equation~\eqref{eq:concrete_score} represents the score function the ExpConcrete distribution i.e. the gradient of the log-density with respect to the data:
\begin{equation}
\label{eq:concrete_score}
     \nabla_{\Tilde{\mathbf{x}}_j} \log p_{\lambda}(\Tilde{\mathbf{x}}; \mathbf{x}) = - \lambda + \lambda K~\sigma(\log \boldsymbol{\mathbf{x}} - \lambda \Tilde{\mathbf{x}})_j
\end{equation}

where $\sigma(\mathbf{z})_i = \frac{e^{z_i}}{ \sum_{k=1}^{K} e^{z_k}}$ is the softmax function. The complete derivation is available in the Appendix.

We can now combine the ideas from Denoising Score Matching and Concrete random variables.
Combining Equation~\eqref{eq:dsm} and~\eqref{eq:concrete_score}, we obtain

\begin{align*}
\label{eq:cdsm}
    J(\theta) &= \mathbb{E}_{q_{\sigma}} \s{ || s_\theta(x) - \nabla_{\Tilde{x}} \log q(\tilde{x} | x) ||^2 }\\
     &= \mathbb{E}_{p_{\lambda}} \s{ || s_\theta(\Tilde{\mathbf{x}}) - \nabla_{\Tilde{\mathbf{x}}} \log p_\lambda(\tilde{\mathbf{x}} | \mathbf{x}) ||^2 } \\
     &= \mathbb{E}_{p_{\lambda}} \s{ || s_\theta(\Tilde{\mathbf{x}}) - \nabla_{\Tilde{\mathbf{x}}} \log p_\lambda(\tilde{\mathbf{x}} ; \boldsymbol{\alpha} = \mathbf{x}) ||^2 } \\
     &= \mathbb{E}_{p_{\lambda}} \s{ || s_\theta(\Tilde{\mathbf{x}}) - (- \lambda + \lambda K~\sigma(\log \mathbf{x} - \lambda \Tilde{\mathbf{x}})) ||^2 } \\
     &= \mathbb{E}_{p_{\lambda}} \s{ || s_\theta(\Tilde{\mathbf{x}}) - \lambda K~\sigma(\log \mathbf{x} - \lambda \Tilde{\mathbf{x}}))  + \lambda ||^2 } \\
     &= \mathbb{E}_{p_{\lambda}} \s{ || s_\theta(\Tilde{\mathbf{x}}) - \lambda K~\sigma(\epsilon)  + \lambda ||^2 }
\end{align*}

Here $\epsilon=\log \mathbf{x} - \lambda \Tilde{\mathbf{x}}$ and can be loosely interpreted as the "logit noise" as it is the difference between the original logit probabilities and the perturbed vector. This formulation is analogous to the simplification utilized by \cite{song2021scorebased, ho2020denoising}. It allows us to train the model to estimate the noise directly as the other variables are known constants. Assume a network $\epsilon_\theta$, that takes the input $\Tilde{\mathbf{x}}$. Following Equation~\eqref{eq:concrete_score}, we can parameterize a score network as $s_\theta(\Tilde{\mathbf{x}})_j = - \lambda + \lambda K~\sigma(\epsilon_\theta(\Tilde{\mathbf{x}}))_j$. We train the network $\epsilon_\theta$ to estimate the noise values $\epsilon$ by the objective below.  

\begin{equation}
\label{eq:train_obj}
    J(\theta)  = \mathbb{E}_{p_{\lambda}} \s{ \lambda^2 K^2 ||( \sigma( \epsilon_\theta(\Tilde{\mathbf{x}}) ) - \sigma(\epsilon) )||^2 }
\end{equation}

Following~\cite{song2019generative}, we can modify our loss to train a Noise Conditioned Score Network with $L$ noise levels i.e. $\lambda \in \br{\lambda_i}_{i=1}^L$:

\begin{equation*}
    J(\theta)  = \sum_{i=0}^{L} \lambda_i^2 K_d^2 ~\mathbb{E}_{\mathbf{x} \sim p_{\text{data}}} \mathbb{E}_{\Tilde{\mathbf{x}} \sim p_{\lambda_i}} \s{ || K ( \sigma( \epsilon_\theta(\Tilde{\mathbf{x}}, \lambda_i) ) - \sigma(\epsilon) )||^2 }
\end{equation*}

Note that our network is now additionally conditioned on the noise level $\lambda$. Finally, our loss objective can be extended to incorporate data with multiple categorical features. For $D$ categories we have:

\begin{equation}
\label{gnsm_obj}
    \sum_{d=0}^{D} \sum_{i=0}^{L} \lambda_i^2 K_d^2 ~\mathbb{E}_{\mathbf{x_d} \sim p_{\text{data}}} \mathbb{E}_{\Tilde{\mathbf{x}}_d \sim p_{\lambda_i}} \s{ || ( \sigma( \epsilon_\theta(\tilde{\mathbf{x}}_d,  \lambda_i) ) - \sigma(\epsilon) )||^2 }
\end{equation}

Here, $K_d$ represents the number of outcomes per category, $x_d$ represents the one-hot vector of length $K_d$, and  $\tilde{\mathbf{x}}_d$ is the continuous, noisy representation of $x_d$ obtained after a Concrete (Gumbel-Softmax) transform.

Thus, we have shown that a Concrete relaxation allows us to model the scores of categorical variables by acting as the noise distribution in the Denoising Score Matching objective. The network will output the scores of the logits representing the categorical feature. Intuitively, these scores are gradients pointing in the direction of the category that maximizes the likelihood of the datapoint.

\subsection{Anomaly Detection}
Once a network is trained with the denoising objective in Equation~\eqref{gnsm_obj}, we 
propose to construct an embedding space to identify anomalies. For a given point $x$, we compute the score estimates for all noise perturbation levels. The resulting vector represents the $L$-dimensional multiscale embedding space:
\begin{equation}\label{eq:norms}
    \eta(x) = \left( \norm{  s_\theta(x, \lambda_1) }_2^2, ... , \norm{  s_\theta(x, \lambda_L) }_2^2 \right)
\end{equation}
where $s_\theta(x, \lambda_i)$ is the noise conditioned score network estimating $\nabla_{x} \log p_{\lambda_i}(x)$.
Following the mechanism laid out by~\cite{mahmood2021multiscale}, we learn ``areas of \emph{concentration}" of the inlier data in the $L$-dimensional embedding space ($\eta(x)$, for $x\sim p$). Concretely, we train a GMM on $\eta(X_{\text{IN}})$, where $X_{\text{IN}}$ represents the set of inliers. At inference time, we first use our score network to compute the embedding space $\eta(x)$ for the test samples and then compute the likelihoods of the embeddings via the trained GMM. The negative of this likelihood is then assumed as the anomaly score for the test samples.

\section{Related Works}

Unsupervised anomaly detection has been tackled by a myriad of methods \cite{pang_deep_2021, ruff_unifying_2021}, with varying success~\cite{han2022adbench}. For the purposes of this work, we primarily focus on unsupervised anomaly detection algorithms that have been successfully applied to tabular data. Every algorithm employs its own assumptions and principles about normality~\cite{aggarwal_introduction_2017}. These principles can be elucidated into three broad detection methodologies: 

 \subsection{Classification-Based}
 
 Here we are referring to one-class objectives, which do not need labeled samples. For example,  One-Class Support Vector Machines (OC-SVMs)~\cite{ocsvm} try to find the tightest hyperplane around the dataset, while Deep Support Vector Data Descriptors (DSVDD~\cite{pmlr-v80-ruff18a}  will compute the minimal hypersphere that encloses the data. Both methods assume that inliers will fall under the margins, and consequently use the distance to the margin boundaries as a score of outlierness. 

\subsection{Distance-Based}

 The assumption here is that outliers will be far away from neighbourhoods of inliers. For example, k-Nearest Neighbours~\cite{peterson2009k} will use the distance to the k-th nearest inlier point as a score of anomaly. Isolation Forests~\cite{isoforest} implicitly use this assumption by computing the number of partitions required to isolate a point. Samples that are far away from their neighbours will thus be isolated with fewer partitions and be labeled as anomalies.

\subsection{Density-based}

These models assume that anomalies are located in low-density regions in the input space. The principle objective is then to learn the density function representative of the typical (training) data. A trained model will be used to assign probabilities to test samples, with low probabilities signifying anomalies. Examples include Gaussian Mixture Models (GMMs)~\cite{reynolds2009gaussian} and their deep learning counter part, Deep Autoencoding Gaussian Mixture Models (DAGMM)~\cite{zong2018deep}. Both models estimate the parameters for a mixture of Gaussians, which are then used to assign likelihoods at inference time. ECOD~\cite{li_ecod_2022} uses a different notion of density and estimates the cumulative distribution function (CDF) for each feature in the data. It then uses the tail probabilities from each learned CDF to designate samples as anomalous.

\hfill

Furthermore, we acknowledge that there are many methods built for anomaly detection in images such as  \cite{schlegl2019f}, \cite{Bergmann_2020_CVPR} and \cite{defard2021padim}. However, they have yet to be successfully applied to tabular data and it is uncertain how to extend them to categorical data types. Conversely, some methods have been built to address \textit{only} categorical data types such as ~\cite{fast_cat} (compression-based) and ~\cite{couplings, pang2021homophily}(frequency-based). Unfortunately, it is difficult to find open-source implementations of these models. It is also non-obvious how to extend them to mixed continuous/discrete features. Our method on the other hand, can handle mixed data types by using the appropriate score matching objective for continuous and categorical features.

\section{Experiments}\label{sec:experiments}

We designed two experiments to evaluate our methodology: a benchmark on tabular data and a vision-based case study. The tabular benchmark will quantitatively assess the performance of GNSM compared to baselines. The case study will demonstrate a real world use case of detecting anomalous segmentation masks.

\subsection{Tabular Benchmark}

We created an experimental testbed with categorical anomaly detection datasets sourced from a publicly available curated repository~\footnote{\scriptsize https://sites.google.com/site/gspangsite/sourcecode/categoricaldata}. Note that for our method, we need to know the number of outcomes for each category, to appropriately compute the softmax over the dimensions. This prevents us from using preprocessed datasets such as those made available by ~\cite{han2022adbench}. It is also why we could not use all the datasets in the curated repository, as some had been pre-binarized.

We first split the datasets into inliers and outliers. Next, we divided the inliers into an 80/10/10 split for train, validation, and test respectively. The validation set is used for early stopping and the checkpoint with the best validation loss is used for inference. The test set is combined with the outliers and used for assessing performance. The categorical features were first converted to one-hot vectors and then passed through a log transform to retrieve logits. We used Standard normalization to normalize any continuous features (only relevant for Census). We compute results over five runs with different seeds.

\begin{table}
    \centering
    \small
    \begin{tabular}{rccc}
      \toprule 
      \bfseries Dataset & \bfseries \# Samples & \bfseries \# Anomalies & \bfseries \# Features \\
      \midrule 
      Bank & 36548 & 4640 & 53 \\
      Census & 280717 & 18568 & 396 (+5 cont.)\\
      Chess & 28029 & 27 & 40 \\
      CMC & 1444 & 29 & 25 \\
      Probe & 60593 & 4166 & 67 \\
      Solar & 1023 & 43 & 41 \\
      U2R & 60593 & 228 & 40 \\
      \bottomrule
    \end{tabular}
    \caption{Statistics of public benchmark datasets. All datasets other than Census are cetegorical only.}\label{tab:data}
\end{table}

We chose four methods to represent baseline performance in lieu of a comprehensive analysis with multiple methods. We were inspired to go this route due to the thorough results reported by ADBench~\cite{han2022adbench}. As the authors describe, no one method outperforms the rest. We picked two representatives for shallow unsupervised methods: Isolation Forests and ECOD. We picked these as they consistently give good performance across different datasets and require little to no hyper parameter tuning. There are much fewer options for unsupervised deep learning methods that have been shown to work on tabular datasets. We chose two models that are popular in this field: DAGMM and DSVDD. Note that these were the only unsupervised deep learning models reported by ~\cite{han2022adbench}. For a fair comparison, we tried to keep the number of model parameters at the same order of magnitude as ours when possible. However, this would often lead to numerical instability for DAGMM. Thus we manually tuned the DAGMM hyperparameters for the best and most stable performance for each dataset.

For our score network, we used a ResNet-like architecture inspired by~\cite{gorishniy2021revisiting}. We replaced \texttt{BatchNorm} layers with \texttt{LayerNorm} and set \texttt{Dropout} to zero. The dimensions of the \texttt{Linear} layers in each block were set to 1024. All activations were set to \texttt{GELU}(~\cite{gelu}) except for the final layer, which was set to \texttt{LeakyReLU}. The number of residual blocks was set to 20. 
To condition the model on the noise scales, we added a noise embedding layer similar to those used in diffusion models \cite{song2021scorebased}. We used the same architecture across all datasets.

Our noise parameter $\lambda$ is a geometric sequence from $\lambda=2$ to $\lambda=20$. Early testing showed that the models gave numerical issues for values lower than 2. For the upper-limit (i.e. the largest noise scale) we chose 20 as it works well to smooth out the probabilities to uniform across all datasets. We set the number of noise scales ($L$) to 20. We compute the score norms on the inliers (train+val) according to Equation~\ref{eq:norms} and train a GMM on the resulting features. The negative likelihoods computed from the GMM are the final outputs of our method.

Extensive architectural details are available in the appendix. We also open sourced our code.~\footnote{GitHub link will be provided upon acceptance}

\subsection{Segmentation Case Study}

The task here is to learn the distribution of ground truth image, segmentation pairs. At inference time, our model will score the outputs of a pretrained segmentation model. Our hypothesis is that our method will correctly detect failure cases i.e. poor segmentations should be ranked higher on the anomaly scale.

While there are many ways to qualitatively define a failure, we will be using popular segmentation metrics (with respect to ground truth masks) as a proxy for performance. We posit that a useful anomaly score should correlate meaningfully with the ground truth segmentation accuracies. 

We compare the anomaly scores against three common segmentation metrics: the Dice similarity coefficient (Dice), the mean surface distance (MSD), and the 95-th percentile Hausdorff distance (95-HD). We chose Dice as it is a popular segmentation metric that measures the overlap between the predicted masks and the ground truth. However, as Dice scores may overestimate performance, it is recommended to additionally report distance based metrics~\cite{valentini2014recommendations, taha2015metrics}. These metrics compute the distance between the surfaces of the predictions and ground truth masks.

We train a convolutional noise conditioned score network on the train-set of the Pascal-VOC segmentation dataset ~\cite{Everingham10}. The input to our model is a pair of images and the one-hot segmentation masks. The model predicts the scores for the segmentation masks only. We chose to use paired data rather than segmentations alone as we want the model to learn whether a segmentation is appropriate for the image. 

As our test subject, we retrieved a pretrained \texttt{DeepLabV3} segmentation model~\cite{chen2017rethinking} from the publicly available PyTorch implementation~\footnote{\tiny https://pytorch.org/vision/stable/models/generated/torchvision.models.segmentation.deeplabv3\_mobilenet\_v3\_large}. This model was trained on a subset oft he COCO dataset~\cite{cocodataset}, using only the 20 categories that are present in the Pascal VOC dataset. We computed the segmentation outputs on the validation set of Pascal VOC. The aforementioned segmentation metrics were then calculated between the predictions and ground truth segmentation masks.

We compare the performance of our method to a convolutional DSVDD. While there may exist specialized segmentation uncertainty estimators, we argue that an unsupervised model provides a more apt comparison. It is reasonable to postulate that both our model and DSVDD could be improved by additionally incorporating segmentation-specific objectives into the training, but that remains outside the scope of this study.

For our score matching network, we adopted the NCSN++ model used by~\cite{song2021scorebased}. The only significant change was in the input/output layers as we are predicting scores over one-hot segmentation masks. For DSVDD, we used the implementation of the original authors~\cite{pmlr-v80-ruff18a}. To keep a fair comparison, we modified the code to use a modern architecture as the backbone (specifically EfficentNetV2~\cite{tan2021efficientnetv2}) and kept the number of parameters similar to our model. Both models were trained to convergence and the best checkpoints (tested over a validation split of the train-set) were used for the analysis.



\section{Results}\label{sec:results}

\begin{table*}[!ht]
    \centering
    \begin{tabular}{c|c|ccccc}
    Dataset & Ano Ratio & IForests       & ECOD           & DAGMM          & DSVDD          & GNSM (Ours) \\
    \midrule
    Bank   & 0.56 & $63.24 \pm~1.74$ & $66.52 \pm~0.57$ & $57.62 \pm~3.36$ & $\mathbf{67.18 \pm~6.94}$ & $65.58 \pm~3.45$ \\
    Census  & 0.40 & $40.64 \pm~2.07$ & $40.96 \pm~0.15$ & $32.90 \pm~5.00$ & $41.18 \pm~3.44$ & $\mathbf{47.79 \pm~2.29}$ \\
    Chess   & 0.01 & $\mathbf{2.31 \pm~1.36}$ & $1.43 \pm~0.05$ & $1.08 \pm~0.44$ & $1.47 \pm~0.54$ & $1.60 \pm~0.68$  \\
    CMC      & 0.17 & $22.72 \pm~1.57$ & $23.79 \pm~1.75$ & $24.99 \pm~5.75$ & $21.99 \pm~6.15$ & $\mathbf{ 25.87 \pm~9.93}$ \\
    Probe    & 0.41 & $92.95 \pm~2.28$ & $95.39 \pm~0.38$ & $66.40 \pm~9.43$ & $89.16 \pm~8.40$ & $\mathbf{ 97.48 \pm~0.62}$ \\
    Solar   & 0.30 & $67.99 \pm~3.48$ & $\mathbf{ 72.23 \pm~0.91}$ & $50.84 \pm~5.19$ & $51.21 \pm~3.94$ & $63.34 \pm~5.41$ \\
    U2R     & 0.04 & $52.74 \pm~12.88$ & $67.84 \pm~1.39$ & $10.06 \pm~6.47$ & $71.17 \pm~24.65$ & $\mathbf{ 82.35 \pm~5.45}$ \\
    \end{tabular}
    \caption{Average Precision across multiple datasets. Higher is better. Each experiment was repeated with 5 different seeds and we report the mean and standard deviations across seeds. IForest and ECOD represent shallow models, while DAGMM and DSVDD represent deep learning models. Ano ratio refers to the ratio of anomalies in the test set.}
    \label{tab:tabular_results}
\end{table*}

\subsection{Performance on Tabular Benchmark}
We report the Average Precision error (AP) which can also be interpreted as the Area Under the Precision Recall curve (AUPR).  Average precision computes the mean precision over all possible detection thresholds. We chose to highlight AP over AUROC as it is a more apt measure for detecting anomalies, where we often have unbalanced classes. Additionally, precision measures the positive predictive value of a classification i.e. the true positive rate. This is a particularly informative measure for anomaly detection algorithms where we are preferentially interested in the performance over one class (outliers) than the other (inliers). We would also like to note that our anomaly ratios in the test set do not correspond with the true anomaly ratio in the original dataset. This is due to our data splitting scheme where our test set is effectively only 10\% the size of inliers.

Table~\ref{tab:tabular_results} shows that our approach performs better or on par with baselines. GNSM achieves significant performance improvements over baselines for Census, Probe and U2R, respectively achieving a 6.61\%, 2.09\%, and 11.18\% improvement over the next best method.

Results for CMC and Bank are less straightforward to interpret as the differences in the models are not statistically significant, made apparent by the large overlap in the standard deviations. This is especially true for deep learning models which have to be optimized via gradient descent. On Solar, ECOD outperforms the rest by a significant margin. However, between deep learning models, GNSM performs notably better. Lastly, every model struggled with Chess, quite possibly due to the exceptionally small anomaly ratio. While Isolation Forests achieves the highest mean, it is uncertain whether the win is statistically significant. One could easily opt in favor of the other methods for this dataset as they achieve more consistent results. Again, between deep learning models, GNSM performs better.

Overall, we observed that the shallow models give more stable and consistent results, with ECOD having the smallest standard deviations on average. Additionally, we note that the reported tabular anomaly detection datasets prove difficult for all algorithms. As such, no one method definitively outperforms the rest; an outcome that coincides with previous findings of~\cite{han2022adbench, pang_deep_2021, ruff_unifying_2021}. However, it is our belief that results in Table~\ref{tab:tabular_results} highlight GNSM as a performant contender in the suite of available algorithms for practitioners looking to detect anomalies in unlabeled data domains.

\subsection{Detecting Segmentation Failures}
\label{seg_results}

We computed the anomaly scores from both GNSM and DSVDD and ranked the images from most to least anomalous. Next, we took the top $K=50$ images (out of 1449) and computed the  Pearson correlation coefficients between the ground truth segmentation metrics and the anomaly scores. We chose the worst ranked images for our analysis as we are interested in the efficacy of these scores for identifying segmentation failures as opposed to assessing the quality of successful segmentations.

\begin{figure}[!htb]
  \centering
  \includegraphics[width=\columnwidth]{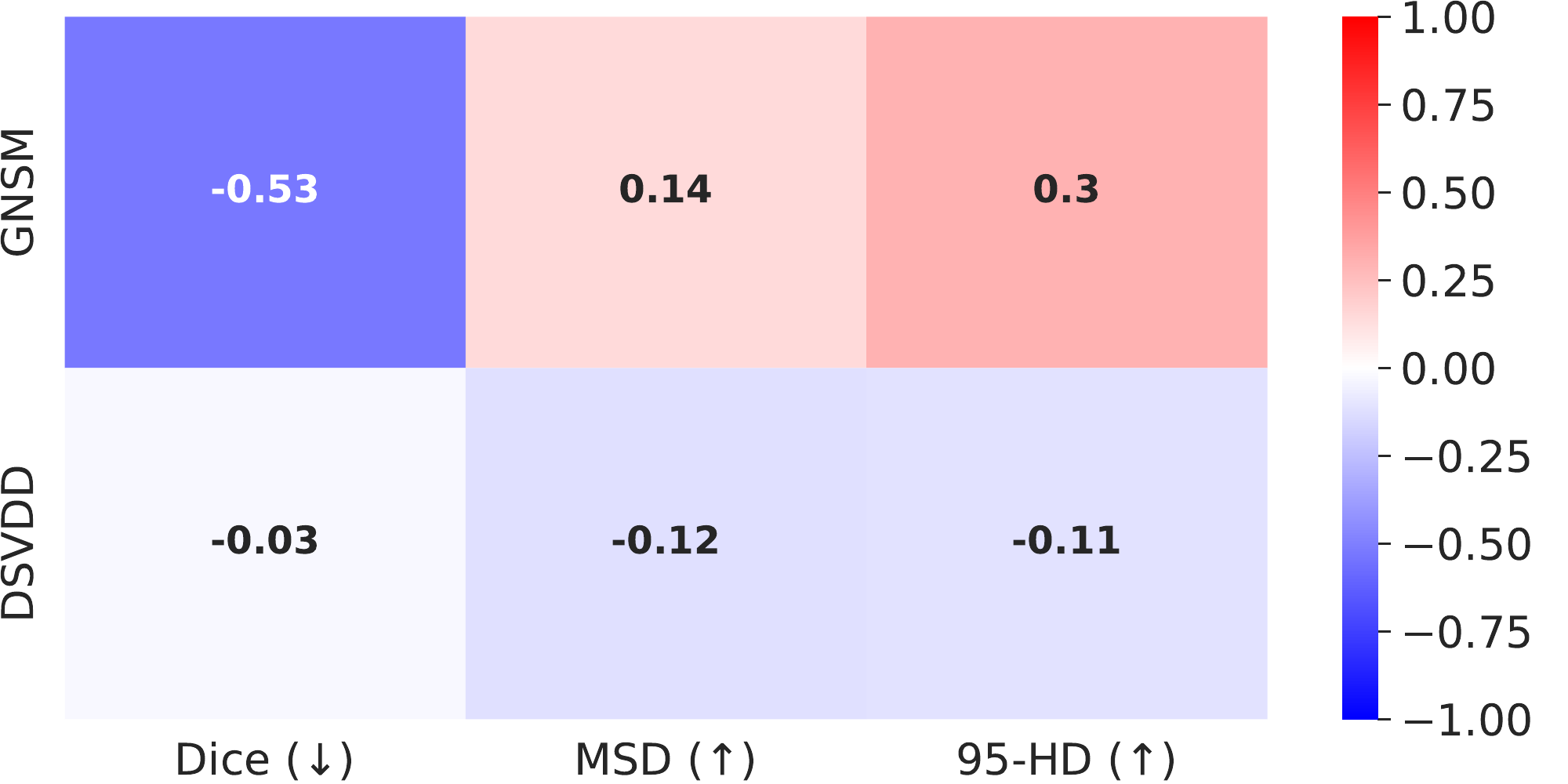}
  \caption{Correlations with segmentation metrics for Top-$K=50$ anomaly scores retrieved from GNSM and Deep SVDD. The arrows next to the metric denote the expected correlation direction. The magnitude of the correlations reflects how well the anomaly scores capture segmentation errors.}\label{fig:corrs}
\end{figure}

Figure~\ref{fig:corrs} shows the correlations between the ground truth segmentation metrics and the anomaly scores from GNSM and DSVDD.  Recall that Dice is a similarity metric while MSD and 95-HD are both distance-based metrics. Therefore, we initially hypothesized that a good anomaly score should correlate negatively with Dice and positively with the distances. Our results show that GNSM correlates strongly in the direction expected. DSVDD on the other hand achieved a poor correlation with Dice and inverse correlations with the distance based metrics. This signifies that our method is meaningfully capturing poor segmentations.

\begin{figure*}\label{fig:samples}
     \begin{subfigure}[t]{0.92\textwidth}
         \centering
        \includegraphics[width=0.95\textwidth, height=4.2in]{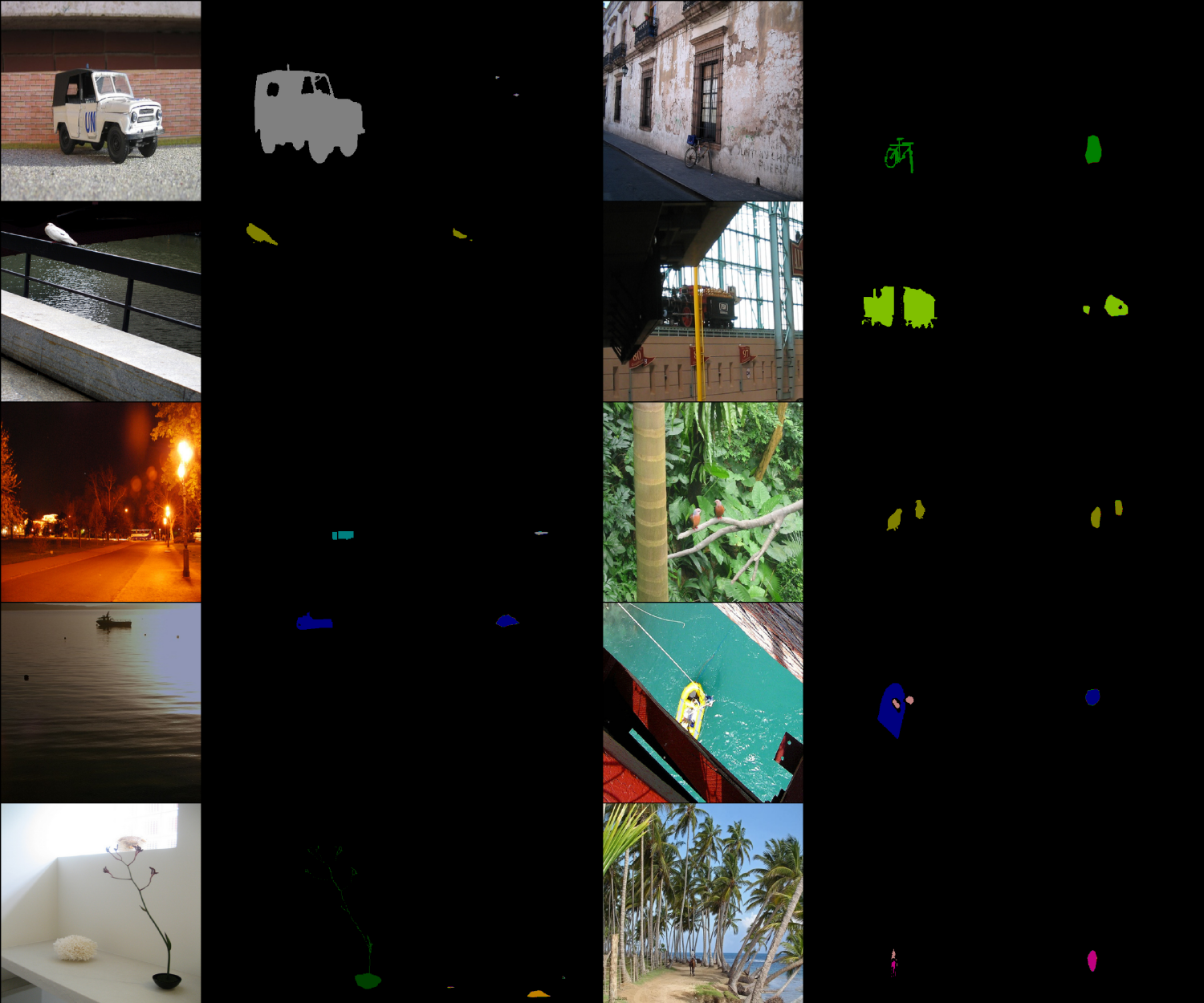}
      \caption{Random samples from Top-K=50 GNSM rankings. Note how the predicted segmentations are either partial/missing or include incorrect classes.}
      \label{fig:msma_samples}
     \end{subfigure}

     \begin{subfigure}[t]{0.92\textwidth}
         \centering
         \includegraphics[width=0.95\textwidth, height=4.2in]{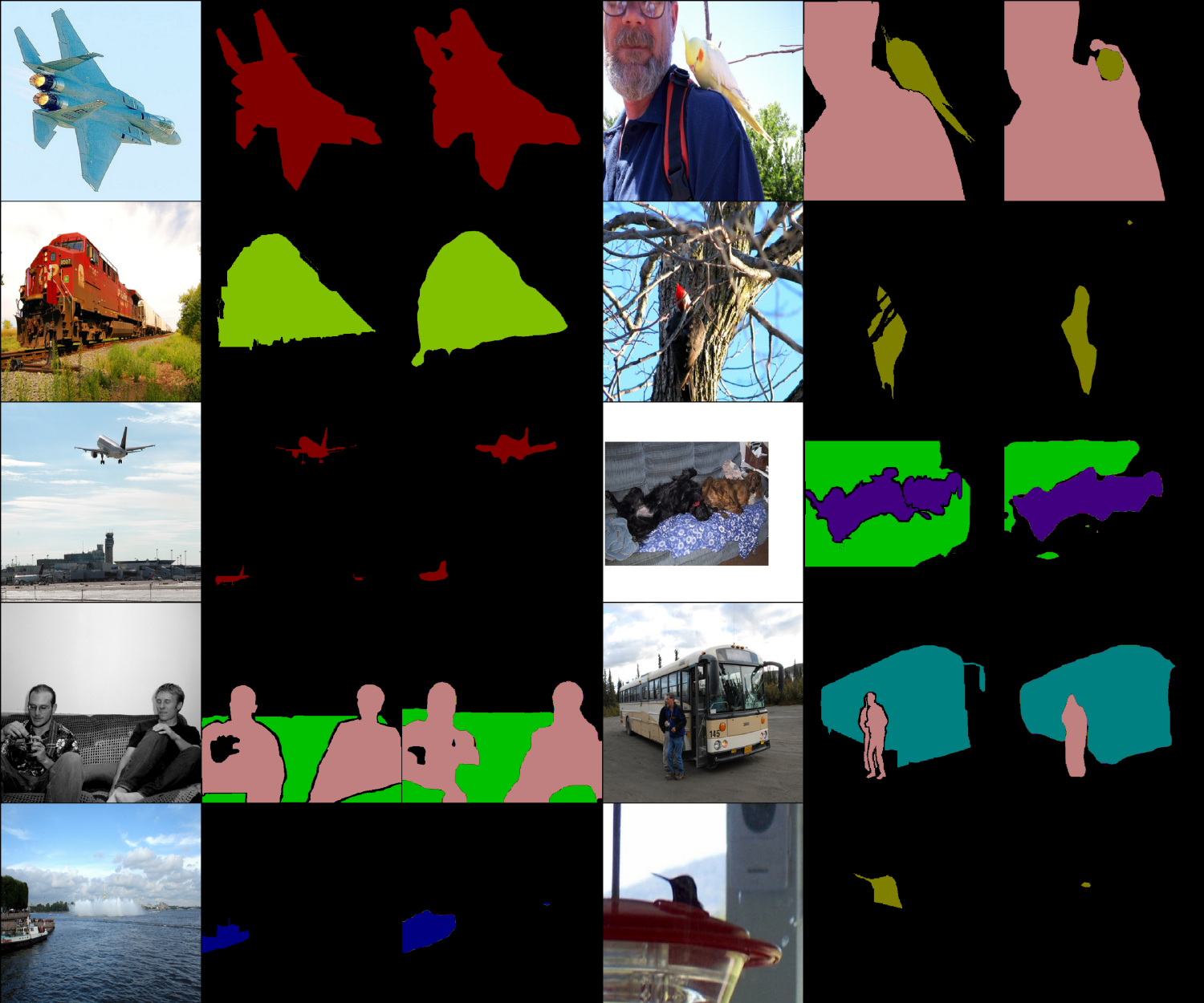}
         \caption{Random samples from Top-K=50 DSVDD rankings. Note how only a few predictions may be considered anomalous.}
         \label{fig:dsvdd_samples}
     \end{subfigure}
\caption{Random samples from Top-K=50 anomaly rankings. The columns (repeated twice) show input image, ground truth segmentations, and model predictions respectively. Different classes are denoted by color.}
\end{figure*}

To qualitatively assess the results of each model and to explain the quantitative results, we visually inspected the worst ranked predictions alongside the groundtruth masks. A subset of the anomalous predictions are plotted in Figure 2. We observe that predictions ranked by GNSM in Figure~\ref{fig:msma_samples} are either complete failures (most of the image is designated the background class) or severe under-segmentations. Predictions ranked by DSVDD in Figure~\ref{fig:dsvdd_samples} generally do not exhibit obvious segmentation failures, with most being reasonable predictions. Please look at the Appendix for all sorted $K=50$ rankings.



\section{Limitations}

Early testing showed that our score networks need to be deep and require more parameters than baselines. Although our proposed network size is performant, we observed a trend of increased performance as the models got deeper and wider. Due to time and resource constraints, we have not thoroughly explored the architecture space. Additionally, our model takes a significant number of iterations to converge. For our experiments, we trained for 1 million iterations, which can take up to a day of training. This is admittedly in contrast to the baselines which may take a few seconds for shallow models and up to a few hours for the deep learning models.

Furthermore, our model explicitly needs to know the number of outcomes per category to appropriately add noise and compute the scores. While we believe this to be a strength of our approach, it does make for an overhead on the user's part. The baselines do not require this additional modeling complexity and are more straightforward to apply. Lastly, our method has hyperparameters pertaining to noise such as the number of scales used and the range of noise levels. While our hyperparameters have been proven to be stable, we posit that additional improvements may be obtained if these were also tuned per dataset.

\section{Conclusion}

In this work we introduced Gumbel Noise Score Matching (GNSM): a novel method for detecting anomalies in categorical data types.
We outline how to compute scores of continuously relaxed categorical data and derive the appropriate training objective based on denoising score matching. Our method can easily be used in conjunction with standard score matching to model both continuous and categorical data.
GNSM achieves competitive performance with respect to baselines on a suite of tabular anomaly detection datasets, attaining significant improvements on certain datasets.
Furthermore, GNSM can easily be extended to images and excels on the real-world task of detecting anomalous segmentations.
Lastly, we believe our novel categorical score matching formulation could be incorporated into generative models. We hope to explore this direction in future work.



\bibliography{main}

\appendix
\onecolumn 

\section{Appendix}

\subsection{Score of ExpConcrete Distribution}
In this section, we compute the score for the ExpConcrete distribution i.e. take the gradient of the log-density with respect to the data. Conveniently, the authors of ~\cite{maddison2017concrete} derived the log-density of an ExpConcrete random variable, which we will be using going forward:

\begin{equation}
    \label{eq:log_pdf}
         \log p_{\alpha, \lambda}(x) = \log((K-1)!) + (K-1) \log \lambda~ + \p{ \sum_{k=1}^{K} \log \alpha_k - \lambda x_k } -
         K\log \sum_{k=1}^{K} e^{(\log \alpha_k - \lambda x_k)}
\end{equation}


Here $x \in \mathbb{N} $ such that $ \log \sum_{k=1}^{K} \exp (x) = 0$. 

Since the first two terms in $\log p_{\alpha, \lambda}(x)$ (Equation~\eqref{eq:log_pdf}) are independent of $x$, we can ignore them and focus on the latter:

\begin{align}
    \nabla_{x_j} \log p_{\alpha, \lambda}(\mathbf{x}) &= 
    \nabla_{x_j} \p{ \sum_{k=1}^{K} \log \alpha_k - \lambda x_k } -
    \nabla_{x_j} \p{K\log \sum_{k=1}^{K} \exp \br{\log \alpha_k - \lambda x_k} } \\
    &= \nabla_{x_j} \p{ - \sum_{k=1}^{K} \lambda x_k } -
   K \p{ \nabla_{x_j} \log \sum_{k=1}^{K} \exp \br{\log \alpha_k - \lambda x_k}} \\
    &= - \lambda -K\frac{\nabla_{x_j} \p{\sum_{k=1}^{K} \exp \br{\log \alpha_k - \lambda x_k}}}{ \sum_{k=1}^{K} \exp \br{\log \alpha_k - \lambda x_k}} \\
    &= - \lambda -K\frac{\exp \br{\log \alpha_j - \lambda x_j} \nabla_{x_j} (\log \alpha_j - \lambda x_j)}{ \sum_{k=1}^{K} \exp \br{\log \alpha_k - \lambda x_k}} \\
    &= - \lambda -K\frac{\exp \br{\log \alpha_j - \lambda x_j} (- \lambda) }{ \sum_{k=1}^{K} \exp \br{\log \alpha_k - \lambda x_k}} \\
    &= - \lambda + \lambda K~\frac{\exp \br{\log \alpha_j - \lambda x_j}}{ \sum_{k=1}^{K} \exp \br{\log \alpha_k - \lambda x_k}}
\end{align}

Note how the last equation can be rewritten as:
\begin{equation}
\label{concrete_score}
     \nabla_{x_j} \log p_{\alpha, \lambda}(\mathbf{x}) = - \lambda + \lambda K~\sigma(\log \boldsymbol{\alpha} - \lambda \mathbf{x})_j
\end{equation}
where $\sigma(\mathbf{z})_i = \frac{e^{z_i}}{ \sum_{k=1}^{K} e^{z_k}}$ is the softmax function.

\subsection{Experiment Details}

\subsubsection{Hyperparameters}
ECOD is hyperparamter free so no tuning was required. Early testing showed that Isolation Forests hyperparameters were stable. Note that we did not use labelled anomalies during hyperparamater tuning and the rest of the deep learning models were tuned on an inlier-only validation set.

\subsubsection*{GNSM Networks}

\begin{equation}
\label{ncsn_obj}
    \sum_{d=0}^{D} \sum_{i=0}^{L} \lambda_i^2 K_d^2 ~\mathbb{E}_{\mathbf{x_d} \sim p_{\text{data}}} \mathbb{E}_{\Tilde{\mathbf{x}}_d \sim p_{\lambda_i}} \s{ || ( \sigma( \epsilon_\theta(\tilde{\mathbf{x}}_d,  \lambda_i) ) - \sigma(\epsilon) )||^2 }
\end{equation}

Observing the loss in Equation~\ref{ncsn_obj}, we see that we are minimizing the difference between two distributions as both inner terms pass through a softmax function. This insight led us to postulate that that we could substitute the mean squared error loss (MSE) for a metric more apt for matching distributions. We therefore ran experiments with KL divergence, which showed faster convergence than MSE. However this result is only empirical and we argue that tuning the optimization algorithm for MSE might gain similar improvements.

\begin{equation}
\label{kl_obj}
    J(\theta)  =  \sum_{d=0}^{D} \sum_{i=0}^{L} \lambda_i^2 K_d^2 ~\mathbb{E}_{\mathbf{x} \sim p_{\text{data}}} \mathbb{E}_{\Tilde{\mathbf{x}} \sim p_{\lambda_i}} \s{ D_{\text{KL}} (\sigma(\epsilon) \parallel  \sigma( \epsilon_\theta(\tilde{\mathbf{x}}_d) ) }
\end{equation}

Most of these details are easily identifiable in our open source code. However, we still provide basic information for posterity. We used the same ResNet-like architecture for all datasets:

\begin{align}
    \texttt{t} &= \texttt{TimeEmbeddingLayer}(\lambda) \\
    \texttt{Net(x, t)} &= \texttt{Head(ResBlock( ... ResBlock( x, t))) } \\
    \texttt{ResBlock(x,t)} &= \texttt{x + Linear(FiLM(x,t))} \\
    \texttt{Head} &= \texttt{Linear(LeakyReLU(LayerNorm(x))}
\end{align}

Note that the \texttt{FiLM} block is taken from ~\cite{perez2018film} and the \texttt{TimeEmbeddingLayer} is the same as used in diffusion models~\cite{song2021scorebased}, using the \texttt{GaussianFourierProjection}.
A simplified implementation of the ResBlock is shown below.

\begin{lstlisting}[language=Python]

class TabResBlockpp(nn.Module):
    def __init__(self, d_in, d_out, time_emb_sz, act="gelu", dropout=0.0):
    
        self.norm = nn.LayerNorm(d_in)
        self.dense_1 = nn.Linear(d_in, d_out)
        self.act = get_act(act)
        self.film = FiLMBlock(time_emb_sz, d_out)
        self.dropout = nn.Dropout(dropout)
        self.dense_2 = nn.Linear(d_out, d_out)

    def forward(self, x, t):
        
        h = self.act(self.norm(x))
        h = self.dense_1(h)
        h = self.film(h, t)
        h = self.dropout(h)
        h = self.dense_2(h)

        return x + h
\end{lstlisting}

For Bank we trained for 2MM iterations while for CMC and Solar, we trained for 600K iterations (as they were significantly smaller datasets). All the other models were trained for 1MM iters. We used the AdamW optimizer with default paramaters. The learning rate was set to $1e-3$ with a cosine decay to $1e-5$ spanning the number of iterations. We also use an Exponential Moving Average of the weights at a decay rate of $0.999$.The base config is shown below.

\begin{lstlisting}[language=Python]
    def get_config():
        config = ml_collections.ConfigDict()
        # training
        config.training = training = ml_collections.ConfigDict()
        training.batch_size = 2048 # Except for CMC and Solar where it was 512
        training.n_steps = 1000001
        training.snapshot_freq = 10000 # Number of iterations for checkpointing
    
        # evaluation
        config.eval = evaluate = ml_collections.ConfigDict()
        evaluate.batch_size = 1024
    
        # data config holds information about the dataset such as number of categories
        config.data = data = ml_collections.ConfigDict()
    
        # default model parameters
        config.model = model = ml_collections.ConfigDict()
        model.name = "tab-resnet"
        model.tau_min = 2.0
        model.tau_max = 20
        ### Only relevant for Census
        model.sigma_min = 1e-1
        model.sigma_max = 1.0
        #####
        model.num_scales = 20
        model.ndims = 1024
        model.time_embedding_size = 128
        model.layers = 20
        model.dropout = 0.0
        model.act = "gelu"
        model.embedding_type = "fourier"
        model.ema_rate = 0.999

        # optimization
        config.optim = optim = ml_collections.ConfigDict()
        optim.weight_decay = 1e-4
        optim.optimizer = "AdamW"
        optim.lr = 1e-3
        optim.beta1 = 0.9
        optim.beta2 = 0.999
        optim.grad_clip = 1.0
        optim.scheduler = "cosine"
\end{lstlisting}

Lastly for MSMA, we train a GMM on the combined train,val set. We run a small grid search over number of components (3,5,7,9) and pick the one with best likelihood.

\subsubsection*{DSVDD}
For Deep SVDD we used the implementation available in the PyOD library~\cite{zhao2019pyod}. Initial testing showed that the autoencoder variant of this model usually performed better. This version adds a reconstruction loss to the one-class objective for increased regularization. The hidden neurons were set to [1024, 512, 256], with the \texttt{swish} activation function. Training was done with the Adam optimizer at default hyperparamters, with learning rate set to 1e-3. We trained for 1000 epochs, with the batch size set to 512.

\subsubsection*{DAGMM}
DAGMM proved to be very difficult to train as most implementations often unexpectedly result in NaNs. In fact the implementation used by~\cite{han2022adbench} never seemed to converge for any dataset. and the loss would not improve no matter how much we tweaked the hyperparameters. We believe the matrix inverse operation during the forward pass to be the culprit for this numerical instability. 

We settled on modifiying a publicly available PyTorch implementation\footnote{https://github.com/lixiangwang/DAGMM-pytorch}. We added the following changes to improve stability and performance:

\begin{itemize}
    \item Added Layer Normalization
    \item Added weight initialization
    \item Included checkpointing and early stopping using val set
    \item GMM parameters converted to double (float64)
\end{itemize}

Furthermore, we hand tuned hyperparameters for each dataset to find the most optimal (stable + performant) setting. Essentially, we tried to start from the same hyperparamters as DSVDD and tweaked until we got a stable model. We also early stopped on the checkpint that gave the best validation loss (tested every epoch). If a NaN was encountered before the first epoch was finished (i.e. before any checkpoint could be saved), we would restart training. The following hyperparameters were used for the final experiments:

\begin{lstlisting}[language=Python]
hyp = {
    "input_dim": input_size,
    "hidden1_dim": 1024,
    "hidden2_dim": 512,
    "hidden3_dim": 256,
    "zc_dim": 2,
    "emb_dim": 128,
    "n_gmm": 2,
    "dropout": 0.5,
    "lambda1": 0.1,
    "lambda2": 0.005,
    "lr": 1e-4,
    "batch_size": 256,
    "epochs": 1000,
    "patience_epochs": 10,
    "checkpoint": "best",
    "return_logits":False,
}

# Taken from KDDCUP-Rev config from original DAGMM paper
# Most other configs are unstable and frequently result in NaNs during training
if config.data.dataset in ["probe", "u2r"]:
    hyp["hidden1_dim"] = 120
    hyp["hidden2_dim"] = 60
    hyp["hidden3_dim"] = 30
    hyp["emb_dim"] = 10
    hyp["n_gmm"] = 4
    hyp["zc_dim"] = 1
    hyp["batch_size"] = 1024
    hyp["return_logits"] = True
    hyp["lr"] = 1e-5

if config.data.dataset == "bank":
    hyp["hidden1_dim"] = 64
    hyp["hidden2_dim"] = 32
    hyp["hidden3_dim"] = 16
    hyp["emb_dim"] = 10
    # hyp["zc_dim"] = 1
    hyp["batch_size"] = 4096
    hyp["lr"] = 1e-5

if config.data.dataset == "census":
    hyp["hidden1_dim"] = 256
    hyp["hidden2_dim"] = 128
    hyp["hidden3_dim"] = 64
    hyp["emb_dim"] = 10
    hyp["lr"] = 1e-5

\end{lstlisting}

\subsection{Extended Results}

Below we show the predicitions ranked poorly by both DSVDD and our method. Images are displayed in order from highest ranking to lowest (displayed left to right).

\begin{figure*}[!tb]
  \centering
  \includegraphics[width=\textwidth]{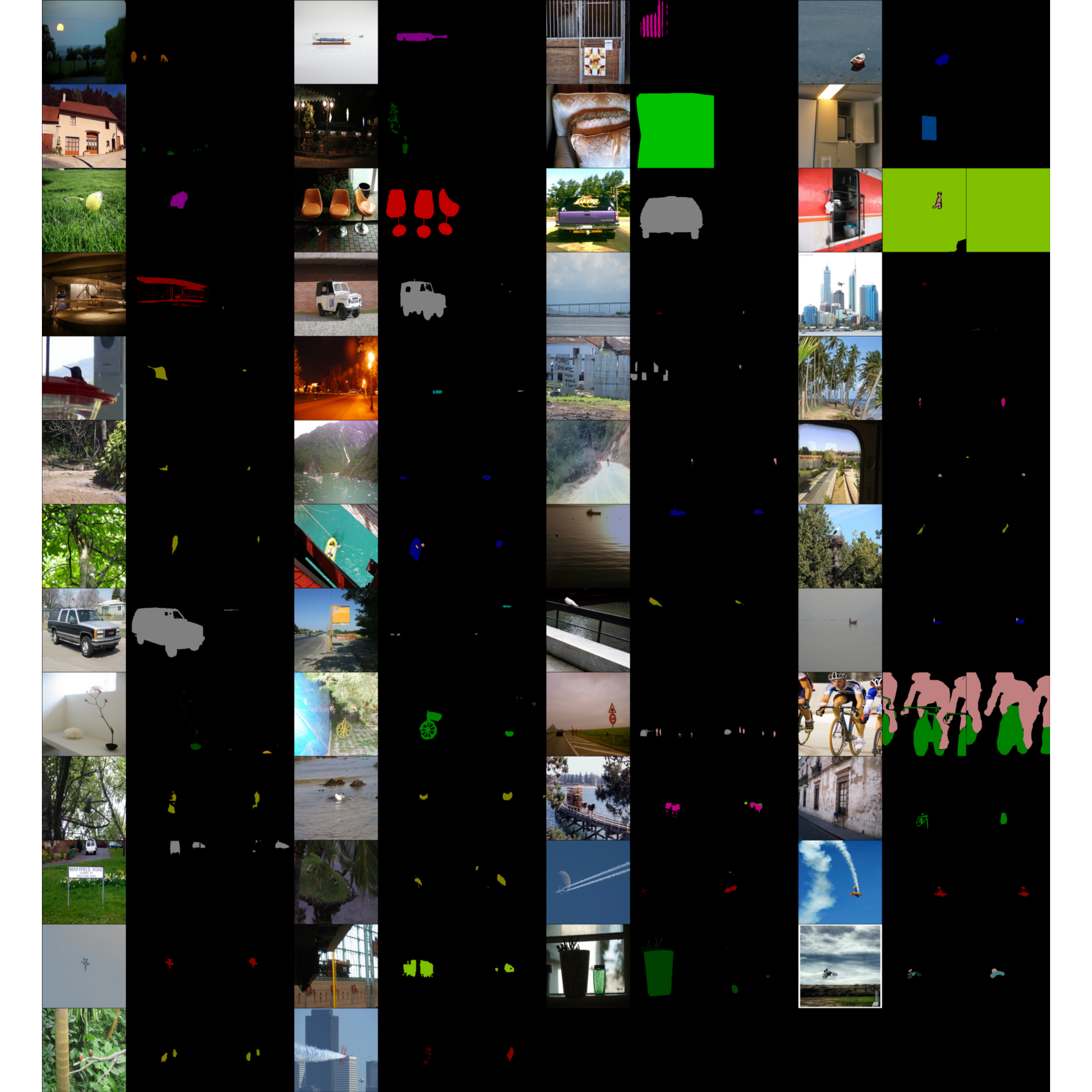}
  \caption{Samples from Top-K=50 GNSM rankings. The columns (repeated) show input image, ground truth segmentations, and model predictions respectively.  }\label{fig:msma_samples_k50}
\end{figure*}

\begin{figure*}[!tb]
  \centering
  \includegraphics[width=\textwidth]{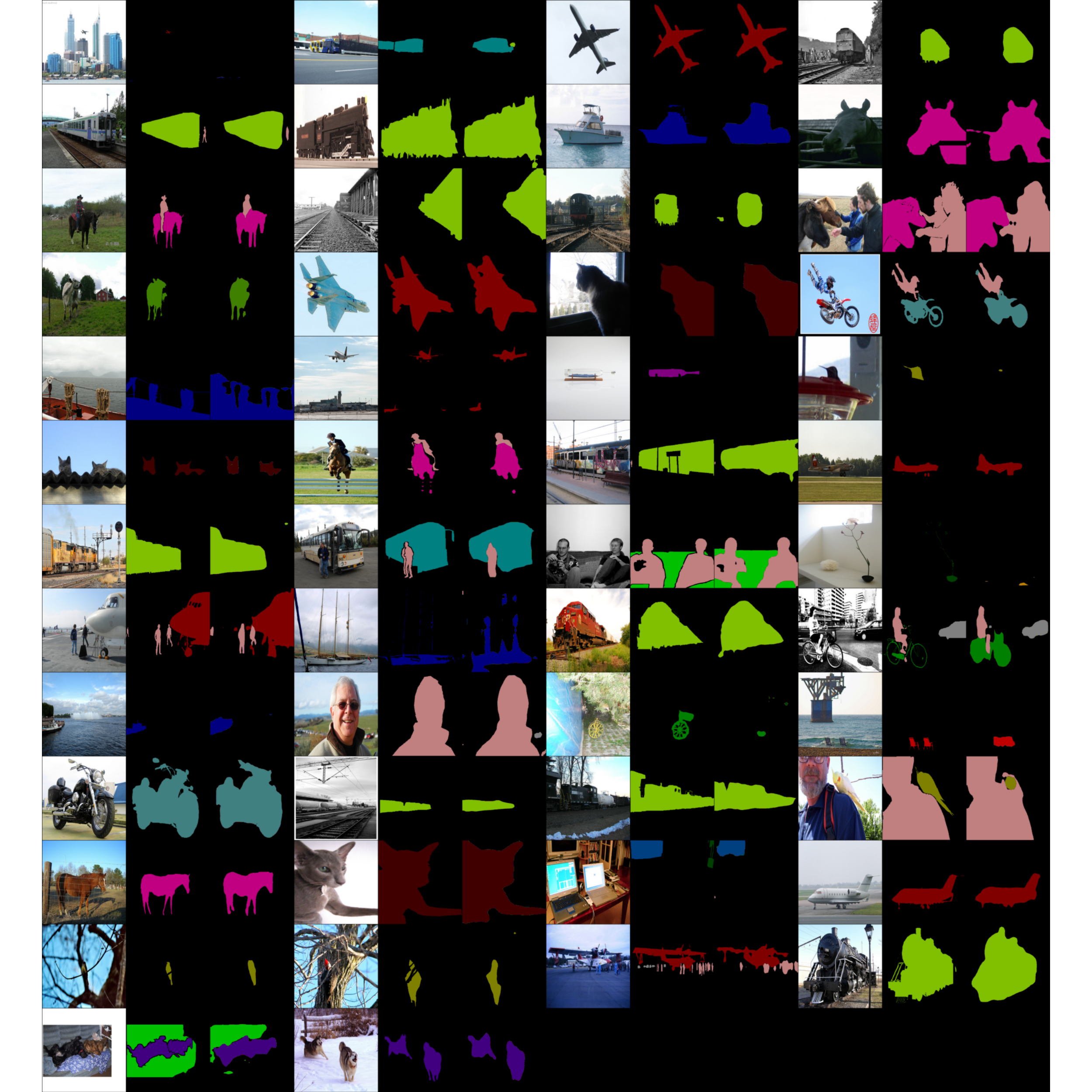}
  \caption{Random samples from Top-K=50 DSVDD rankings. The columns (repeated) show input image, ground truth segmentations, and model predictions respectively.  }\label{fig:dsvdd_samples_k50}
\end{figure*}

\end{document}